\documentclass[letterpaper]{article} 
\usepackage{aaai24}  

\usepackage{times}  
\usepackage{helvet}  
\usepackage{courier}  
\usepackage[hyphens]{url}  
\usepackage{graphicx} 
\usepackage{natbib}  
\usepackage{caption} 
\usepackage{algorithm}
\usepackage{algorithmic}

\usepackage{multirow}
\usepackage{multicol}
\usepackage{booktabs}

\usepackage{subcaption}
\usepackage{pifont}

\usepackage[table]{xcolor}

\usepackage{newfloat}
\usepackage{listings}
\DeclareCaptionStyle{ruled}{labelfont=normalfont,labelsep=colon,strut=off} 
\floatstyle{ruled}
\newfloat{listing}{tb}{lst}{}
\floatname{listing}{Listing}
\nocopyright

\title{Patch Is Not All You Need}
\author{
    Changzhen Li\textsuperscript{\rm 1,\rm 2,\rm 3},
    Jie Zhang\textsuperscript{\rm 1,\rm 2},
    Yang Wei,
    Zhilong Ji\textsuperscript{\rm 4},
    Jinfeng Bai\textsuperscript{\rm 4},
    Shiguang Shan\textsuperscript{\rm 1,\rm 2,\rm 3},
}
\affiliations{
    \textsuperscript{\rm 1}Key Lab of Intelligent Information Processing of Chinese Academy of Sciences, Institute of Computing Technology\\
    \textsuperscript{\rm 2}University of Chinese Academy of Sciences\\
    \textsuperscript{\rm 3}Hangzhou lnstitute for Advanced Study, UCAS, school of Intelligent Scienceand Technology\\
    \textsuperscript{\rm 4}Tomorrow Advancing Life\\

}

\usepackage{bibentry}

\begin{document}

\maketitle

\begin{abstract}
Vision Transformers have achieved great success in computer visions, delivering exceptional performance across various tasks. However, their inherent reliance on sequential input enforces the manual partitioning of images into patch sequences, which disrupts the image's inherent structural and semantic continuity. To handle this, we propose a novel Pattern Transformer (Patternformer) to adaptively convert images to pattern sequences for Transformer input. Specifically, we employ the Convolutional Neural Network to extract various patterns from the input image, with each channel representing a unique pattern that is fed into the succeeding Transformer as a visual token. By enabling the network to optimize these patterns, each pattern concentrates on its local region of interest, thereby preserving its intrinsic structural and semantic information. Only employing the vanilla ResNet and Transformer, we have accomplished state-of-the-art performance on CIFAR-10 and CIFAR-100, and have achieved competitive results on ImageNet.

\end{abstract}

\section{Introduction}
In natural language processing, each word represents a unique concept, and sentences are composed from sequences of these abstract words. Transformer model, by treating each word as a token, leverages the self-attention mechanism to model global contextual relationships, providing a fundamental advantage in processing textual sequences \cite{vaswani2017attention,devlin2018bert,brown2020language}. Inspired by this, Vision Transformer splits images into sequences of patches during the preprocessing phase, which are then treated as visual ``words'' and fed into the Transformer network \cite{dosovitskiy2020image}. However, despite the significant success of Transformers in vision tasks, the tokenization procedure introduces several challenges:
(1) Collapse the structural and semantic information. Unlike natural language where each word constitutes an independent entity adhering to a relatively fixed sentence structure (e.g., Subject + Verb + Object), the structural information in an image is complex and irregular. The manual division disrupts the innate structure of objects within the image, and meanwhile, the image's variance (e.g., scale, rotation) results in inconsistent semantic sequences for identical objects.
(2) Images inherently possess a high degree of redundancy compared to the concise concepts of text. The procedure of manually partitioning images produces a fixed number of patches, many of which are redundant and uninformative for the target class. These redundant patches are not only fail to contribute beneficially but also impose a computational burden on expensive Transformer.

\begin{figure}[t]
\centering
\includegraphics[width=0.9\columnwidth]{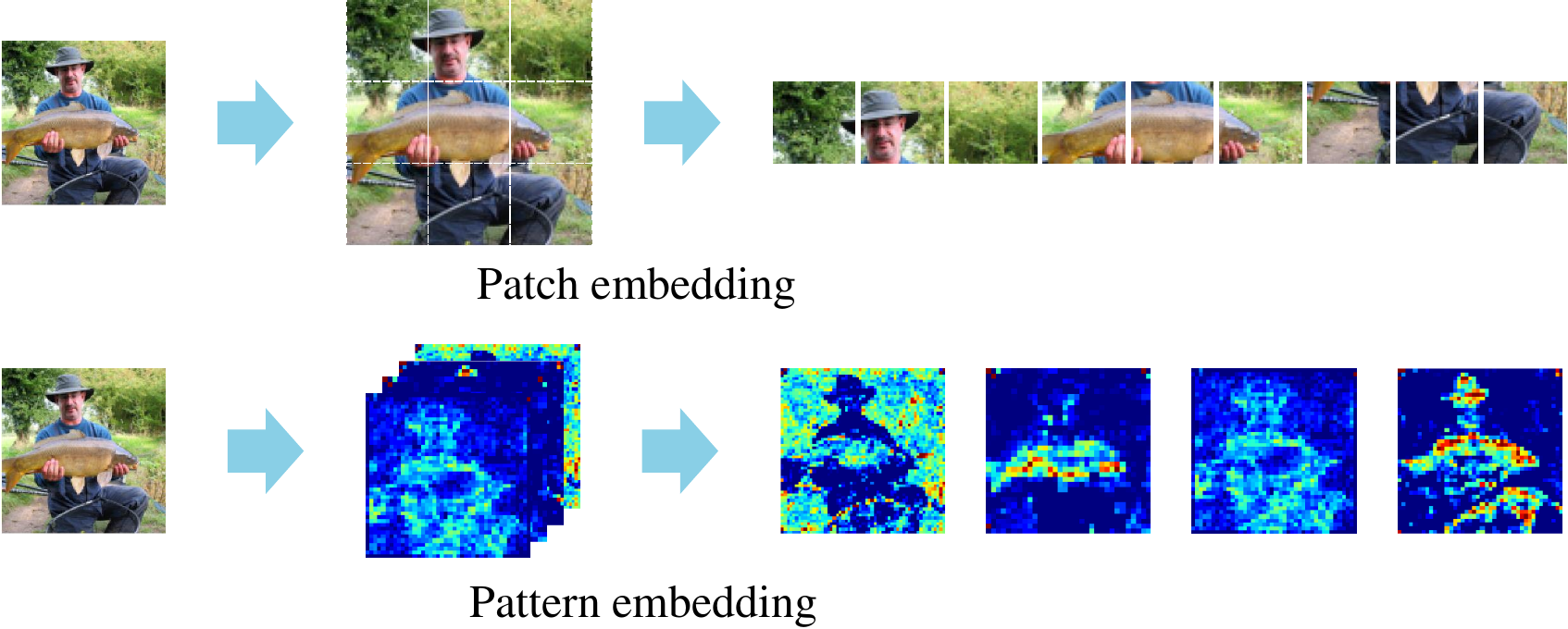} 
\caption{Patches are local spatial features, while Patterns are global semantic features.}
\label{intro}
\vspace{-3pt}
\end{figure}

\begin{figure*}[t]
\centering
\includegraphics[width=0.8\textwidth]{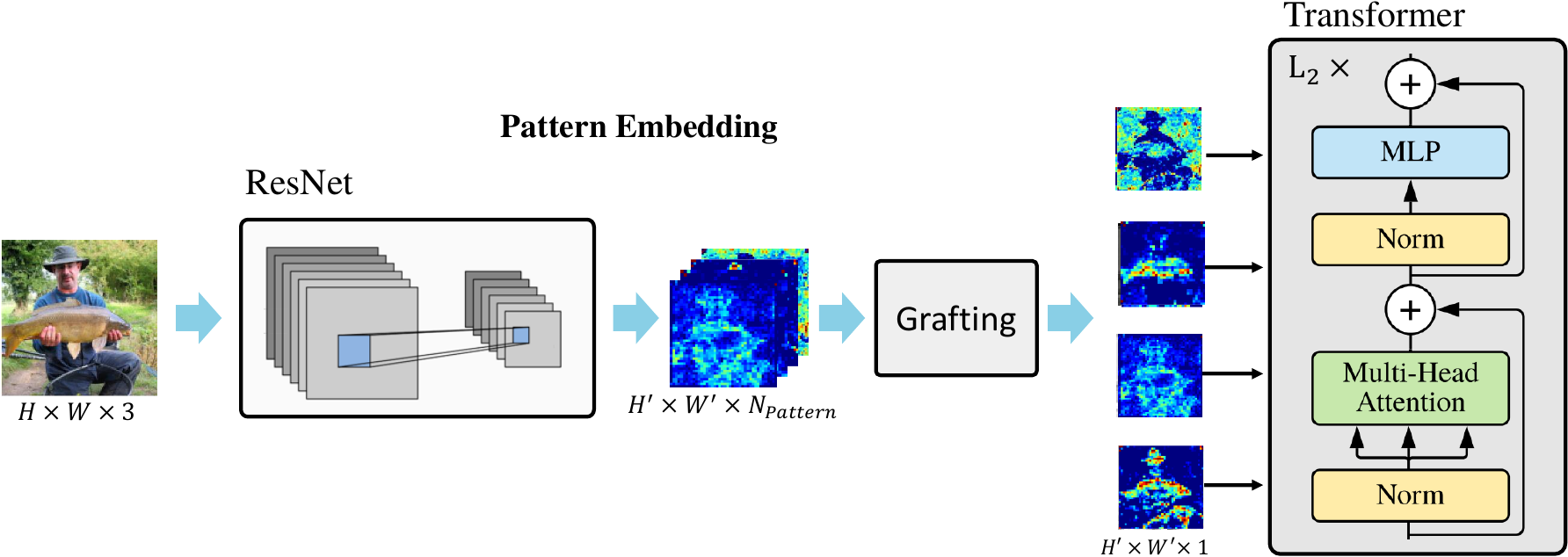}
\caption{Pipeline. First, capture local semantic patterns by employing the vanilla ResNet, and then model the global context by employing the vanilla ViT, which leverages the strengths of both CNNs and ViTs.}
\label{pipeline}
\vspace{-3pt}
\end{figure*}

Despite the inherent challenges involved in converting images to sequences, this critical issue has yet to find an effective solution, with most methods persisting with the patchification following ViT \cite{dosovitskiy2020image}. A cohort of researchers have sought to improve the patch embedding methodology. For example, CeiT \cite{yuan2021incorporating} substitutes the initial linear projection with a convolutional stem, thereby combining the strengths of CNNs in extracting local features. DPT \cellcolor{chen2021dpt} introduces Deformable Patch to adaptively divide patches based on varying positions and scales. TokenLearner \cite{ryoo2021tokenlearner} focuses on learning eight crucial visual tokens to model images or videos, however, these tokens, once subjected to global pooling, completely discard spatial information. Another faction strives to mitigate the negative impact of patchification. FlexiViT \cite{beyer2023flexivit} recognizes the limitation of a fixed patch size and utilizes flexible patch size to design trade-off models. Some suggest removing random (as in MAE \cite{he2022masked}) or unimportant (as in A-vit \cite{yin2022vit}) tokens. Although these methods improve the process of converting images to sequences and tackle the issue of the fixed patch size and redundant patches, they still fail to preserve the structural and semantic information within image objects as long as the patch partitioning remains in use.

To address these issues, we propose a novel Pattern Transformer (Patternformer) to adaptively converting images to pattern sequences. These sequences serve as the inputs of Transformer, thereby eliminating complications introduced by manual patchification. A pattern typically denotes a specific structure or an underlying rule within an image, such as faces, buildings, vehicles, or parts of some objects. The interpretability of neural networks reveals that certain channels within specific layers capture distinct patterns of local regions \cite{bau2017network}. Based on this, we leverage a convolutional neural network to extract various patterns from the image, with each channel representing a pattern that is fed into the Transformer as a visual token. Converting images into sequences by patterns eliminates the need to manually divide the image into rigid patches based on experiential knowledge. It captures the local region of interest in accordance with network optimization, thus preserving its structural and semantic information. Moreover, the process of pattern embedding does not involve image size and patch size, resulting in a flexible and variable sequence length. This flexibility improves the efficiency of the Transformer's modeling capabilities.

Overall, we employ CNNs to adeptly capture local patterns and Transformer to model the global context, thus effectively capitalizing on the inherent advantages of both CNNs and Transformer. We conducted extensive experiments and accomplished state-of-the-art performance on CIFAR-10 and CIFAR-100, and achieved competitive results on ImageNet.

\section{Related Work}
\subsubsection{Convolutional Neural Networks}
Convolutional Neural Networks (CNNs) have emerged as a dominant paradigm in image classification for the last decade. Pioneering architectures such as AlexNet, VGGNet, Inceptions, ResNet, DenseNet continued to push the boundaries of image classification accuracy through deeper and wider structures \cite{krizhevsky2012imagenet,simonyan2014very,szegedy2015going,he2016deep,huang2017densely}. Subsequent models like MobileNet, ShuffleNet, EfficientNet, and RegNet focused on a better trade-off of accuracy and efficiency \cite{howard2017mobilenets,tan2019efficientnet,radosavovic2020designing}. Despite their success, CNNs inherently struggle to capture long-range dependencies, a capability that can be crucial for comprehending complex scenes.

\subsubsection{Vision Transformers}
Vision Transformers have recently emerged as a promising alternative to CNNs, inspired by the success of Transformers in Natural Language Processing (NLP). ViT \cite{dosovitskiy2020image} was the first to apply a pure transformer to the sequences of image patches for image classification. Following this, numerous strategies have been proposed to further improve vision transformer. e.g., data efficient (DeiT) \cite{touvron2021training}, 
modeling local feature (Swin Transformers, TNT, Shuffle Transformer, RegionViT) \cite{liu2021swin,han2021transformer,huang2021shuffle,chen2021regionvit}, improving self-attention layer (DeepViT, KVT, XCiT ) \cite{zhou2021deepvit,wang2022kvt,ali2021xcit}, pyramid architecture(PVT, PiT) \cite{wang2021pyramid,heo2021rethinking}, Neural architecture search (ViTAS) \cite{su2022vitas}. However, Vision Transformers require massive training data, and lack the ability to keep the innate structure of objects under various transformations like scale and rotation.

\subsubsection{Hybrid Vision Transformers}
To combine the strengths of both CNNs and ViTs, Hybrid Vision Transformers have been proposed. 
CvT \cite{wu2021cvt}, CMT \cite{guo2022cmt}, CeiT \cite{yuan2021incorporating}, and LocalViT \cite{li2021localvit} all bring convolutions into transformers blocks to promote the correlation among neighboring tokens.
LeViT \cite{graham2021levit} and BoTNet \cite{srinivas2021bottleneck} resort to a different technology roadmap, replacing former transformers with convolutions, thereby proposing a hybrid neural network for faster inference.

Our Pattern Transformer is also a hybrid Vision Transformers, but we regard each pattern as visual tokens to feed into the Transformer, which is totally different from existing vision transformers.

\begin{figure*}[!ht]
    \centering
    \begin{subfigure}{0.26\textwidth}
        \centering
        \includegraphics[height=0.86\linewidth]{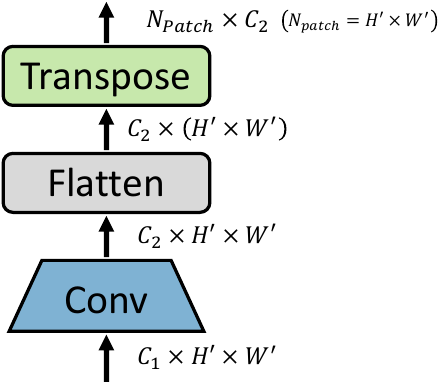}
        \caption{vanilla ViT}
        \label{pattern1}
    \end{subfigure}
    \begin{subfigure}{0.26\textwidth}
        \centering
        \includegraphics[height=0.86\linewidth]{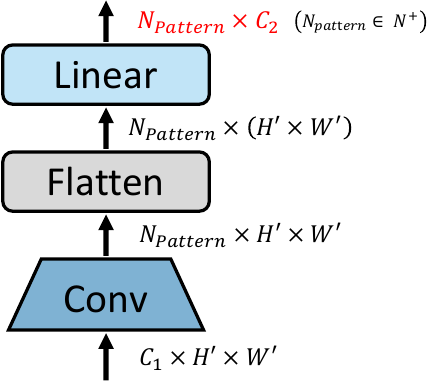}
        \caption{our Pattern Transformer}
        \label{pattern2}
    \end{subfigure}
    \begin{subfigure}{0.26\textwidth}
        \centering
        \includegraphics[height=0.86\linewidth]{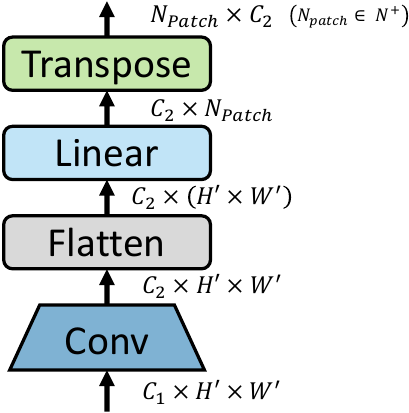}
        \caption{our Patch Transformer}
        \label{pattern3}
    \end{subfigure}
    \caption{The Grafting combines the ResNet and ViT}
    \label{pattern}
\end{figure*}

\section{Method}
\label{san}
We illustrate the overall diagram of the Pattern Transformer architecture in Figure \ref{pipeline}, comprising two primary components. 
First, capture local semantic patterns by employing the vanilla ResNet \cite{he2016deep}, and then model the global context by employing the vanilla ViT \cite{dosovitskiy2020image}, which leverages the combined strengths of both CNNs and ViTs.

\subsubsection{Revisiting Vision Transformer}
The standard Transformer processes a 1D sequence of patch embeddings as input. To accommodate 2D images, ViT reshapes the image $x \in R^{H \times W \times 3}$ into a sequence of flattened 2D patches $x_{patch} \in R^{N_{patch} \times (P^2 \cdot C)}$, where $(H, W)$ represents the resolution of the original image, $C$ represents the number of channels, $(P, P)$ denotes the resolution of each image patch, and $N_{patch} = HW/P^2$ is the resulting number of patches, which also serves as the effective input sequence length for the Transformer. 
In practice, ViT splits each image into patches with fixed size and positions by employing single or multiple convolution layers, and these patches possess a feature size ${1 \times 1 \times C}$ and a receptive field ${P \times P}$. However, this process introduces two limitations: (1) The patch with a $P \times P$ receptive field is hard to be an independent entity, which collapses the structural and semantic information of the original image. (2) The sequence length $N$ is constrained by the values of the image resolution $(H, W)$ and patch size $P$, and moreover, all $N$ patches are treated equally, even those lacking sufficient discriminative information.

\subsubsection{Pattern Embedding}
Our intuition is that an image can be summarized by a sequence of visual ``words''. This contrasts convolutions, which employ hundreds of filters to detect all possible concepts of image content. Inspired by this, we introduce Pattern Embedding to convert the image $x \in R^{H \times W \times 3}$ into a compact sequence of visual words ${x_{pattern} \in R^{N_{pattern} \times ((H^{'}, W^{'}) \cdot 1)}}$, where $(H^{'}, W^{'})$ represents the resolution of each pattern, ``1'' represents a certain channel, $N_{pattern}$ is the resulting number of patterns. The pattern with $H^{'} \times W^{'} \times 1$ is detected by a specific filter, representing an independent semantic concept. And due to its global receptive field ${H \times W}$, each pattern does not disrupt the structural and semantic information. Furthermore, The sequence length $N_{pattern}$ is a hyperparameter $(N_{pattern} < N_{patch})$, which is smaller than the number of patches to ensure compact and efficient information. To achieve this, we design a heavy and flexible Tokenizer to meet all above.

\subsubsection{Heavy Tokenizer}
\label{heavy}

We utilize the entire ResNet as a heavy Tokenizer to detect all potential image content concepts, for instance, patterns which are represented as each channel of output feature maps. Different from existing works, we view the entire pattern with a global receptive field as visual words, which is further fed to Transformer for global context learning.

\textbf{Building Block and Width}. BasicBlock (Basic) employs a stack of two 3x3 convolutions with a fixed width of 64. Bottleneck (Bottle) employs a stack of three convolutions with an optional width, where 1x1 layers adjust dimensions and the 3x3 layer acts as a reduced-dimension bottleneck. We utilize the Bottleneck with an optional width to balance speed and accuracy.

\textbf{Changing stage ratio}. The original design of the computation distribution across stages that makes ResNet more efficient than VGGNet is to apply strong resolution reductions with a relatively small computation budget in its first two stages. We further optimize the architecture for similar reasons and adjust the number of blocks in each stage from [3,4,6,3] in ResNet50 to [1,1,6,3]. Unlike existing designs, e.g., the stage ratio 1:1:9:1 in Swin Transformer, a heavy ``stage4'' greatly helps in extracting rich patterns.

\subsubsection{Flexible Tokenizer}
The vanilla ViT \cite{dosovitskiy2020image} has experimented with a hybrid vision transformer by stacking the Transformer above ResNet, and the patch embedding is applied to patches extracted from ResNet activation maps as illustrated in Figure \ref{pattern1}. The output activation map by ResNet is $x_1 \in {C_1 \times H^{'} \times W^{'}}$, where $H^{'}=H/P,W^{'}=W/P$, P is the downsampling size of ResNet. 
\begin{equation}
    x_2 = Transpose(Flatten(Conv(x_1)))
\end{equation}
The input feature map of ViT is $x_2 \in N_{patch} \times C_2, N_{patch} = (H^{'} \times W^{'})$, where the length of patch sequence is constrained by $H^{'}$ and $W^{'}$.

\textbf{Removing Transpose Operation}
Traditional Transformers utilize self-attention to capture long-range dependencies among patches, thereby necessitating a Transpose operation to align the $N_{patch}$ with the sequence length in Transformer during the Grafting process from ResNet to ViT. In contrast, our Pattern Transformer exactly removes the Transpose operation to align channels with the sequence length in Transformer, resulting in the computation of self-attention among patterns.

\begin{table*}[!ht]
\centering
\small
    \begin{tabular}{l|ccc|ccccc|cc}
    \toprule
    \multirow{2}{*}{Model} & \multicolumn{3}{c|}{Resnet} & \multicolumn{5}{c|}{Transformer}        & \multirow{2}{*}{Params} & \multirow{2}{*}{GFLOPs} \\
                           & Block   & Width   & Stages  & Tokens$^*$ & Embedding & Depth & Heads & Mlp ratio &  &               \\ \hline\hline
    Res$34$ - ViT$S$    & basic   &64       & [3,4,6,3]    & 128   & 384   & 12    & 6       & 4          &42.8M    &6.6                  \\
    Res$34$ - ViT$B$    & basic   &64       & [3,4,6,3]    & 128   & 768   & 12    &12       & 4          &106.6M   &15.0                   \\
    Res$50$ - ViT$S$    & bottle  &64       & [3,4,6,3]    & 128   & 384   & 12    & 6       & 4          &45.2M    &7.0                   \\
    Res$50$ - ViT$B$    & bottle  &64       & [3,4,6,3]    & 128   & 768   & 12    &12       & 4          &109.0M   &15.4                   \\ 

    Efficient-$T$    & bottle  &32       & [1,1,6,3]    & 64   & 192   & 6    &6       & 2                  &11.9M   &1.6                   \\ 
    Efficient-$S$    & bottle  &64       & [1,1,6,3]    & 64   & 384   & 6    &6       & 2                  &29.8M   &3.5                   \\ 
    Efficient-$B$    & bottle  &96       & [1,1,6,3]    & 64   & 576   & 6    &6       & 2                  &56.7M   &6.3                  \\ 
    \bottomrule
    \end{tabular}
\caption{Variants of our Pattern Transformer architecture. Tokens number represents the length of pattern sequence \\ in Transformer} 
\label{model}
\end{table*}

\textbf{Introducing Extra Linear Transformation}
Pattern Transformer computes long-range dependencies among patterns, with each pattern $H^{'} \times W^{'} \times 1$ encompassing a global receptive field $H \times W$. As depicted in Figure \ref{pattern2}, we utilize a convolution to convert $C_1$ to $N_{patch}$ for flexible adjustment of the pattern sequence length. Similarly, we introduce an extra Linear layer to convert $H^{'} \times W^{'}$ to $C_2$ for flexible adjustment of the dimension of patterns.
\begin{equation}
    x_2 = Linear(Flatten(Conv(x_1)))
\end{equation}
Consequently, this Grafting process facilitates the combination between any ResNet and ViT models, without consideration of their inherent feature size.

As a byproduct, we also introduce an extra Linear layer into the vanilla ViT, thereby constructing a flexible Patch Transformer as shown in Figure \ref{pattern3}. 
\begin{equation}
    x_2 = Transpose(Linear(Flatten(Conv(x_1))))
\end{equation}
For instance, the R50+ViT-B architecture in ViT \cite{dosovitskiy2020image}, constrained by the upstream ResNet output features size, used only 49 patches for the subsequent Transformer. Employing our Patch Transformer, we convert 49 patches up to 128 patches, significantly enhancing network performance.

\subsection{Light Transformer}
The Transformer layer consists of two alternating sub-layers of multi-head self-attention (MSA) and MLP blocks. Residual connections are applied around each sub-layer, followed by a layer normalization (LN). To facilitate residual connections, all sub-layers produce outputs with the same feature dimension. Given that heavy Resnet can extract high-level semantic patterns, we employ a relatively light Transformer to model global dependencies.

\textbf{Reducing Transformer Tokens}
Token number in the vanilla ViT is determined by the image resolution and patch size, for instance, the length of the patch sequence is 196 in the ViT-B architecture. However, we utilize Pattern Transformer with an optional width, such as 64, to balance between speed and accuracy.

\textbf{Reducing Depths and Heads}
The standard ViT contains 12 Transformer layers, with 12 parallel self-attention operations, called ``heads'', in each layer. To reduce the computational cost, we utilize 6 Transformer layers with 6 heads.

\textbf{Reducing the MLP block ratio}
The MLP block is a linear layer that increases the embedding dimension by a factor 4, applies a GELU non-linearity, and reduces it back with another linear layer to the original embedding’s dimension. To reduce the computational cost again, we decrease the expansion factor of the linear layer from 4 to 2. 

\subsection{Pattern Transformer Family}
Pattern Transformer models can spawn a range of speed-accuracy trade-offs by altering the feature size in ResNet and ViT. Table \ref{model} provides an overview of the models considered in our paper. For example, except for employing fewer tokens for the delicate Grafting, the parameters of Res34-ViT$B$ adopt identical parameters as ResNet34 and ViT-B.

\begin{table*}[!ht]
\centering
    \begin{tabular}{lccc|cc}
        \toprule
        Method & Arch. & Params & GFLOPs & CIFAR-10 & CIFAR-100 \\ \hline\hline
        ResNet18 \cite{choi2022tokenmixup}    & \multirow{7}{*}{CNN}   & 11.2M     & -  & 90.27   & 63.41         \\
        ResNet50 \cite{choi2022tokenmixup}    &    & 23.5M     & -  & 90.60   & 61.68         \\
        WRN28-10 \cite{zagoruyko2016wide}       &   & 36.5M    & 5.2  & 96.00   & 80.75            \\
        ResNeXt-29, 8×64d \cite{xie2017aggregated}    &   & 34.4M    & 5.4  & 96.35 & 82.23            \\
        SENet-29 \cite{hu2018squeeze}       &   & 35.0M    & 5.4  & 96.32 & 82.22            \\
        SKNet-29 \cite{li2019selective}       &   & 27.7M    & 4.2  & 96.53  & 82.67            \\
        \hline
        DeiT-S  \cite{touvron2021training}     & \multirow{5}{*}{Transformer}           &22.1M   & 4.3  &92.44 &69.78      \\
        DeiT-B  \cite{touvron2021training}     &            &86.6M   & 16.9  &92.41 &70.49      \\
        PVT-S   \cite{wang2021pyramid}        &            &24.5M   &3.8  &92.34 &69.79      \\
        Swin-S \cite{liu2021swin}         &            &50.0M   & -  &94.17 &77.01      \\
        Swin-B \cite{liu2021swin}         &            &88.0M   & -  &94.55   &78.45      \\
        \hline
        CCT-7/3x1 \cite{hassani2021escaping}      & \multirow{4}{*}{Transformer}           &3.8M   &1.2  &97.48 &82.72      \\
        CCT-7/3x1 + TokenMixup \cite{choi2022tokenmixup}      &            &3.8M   &1.0 &97.75  &83.57      \\
        NesT-S \cite{zhang2022nested}         &            &23.4M   &6.6 &96.97  &81.70      \\
        NesT-B \cite{zhang2022nested}         &            &90.1M   &26.5 &97.20  &82.56      \\
        \hline
        Patternformer (Res34-ViT$S$)              & \multirow{7}{*}{Transformer}        &42.8M    &6.6   &97.72  &83.27     \\
        Patternformer (Res34-ViT$B$)              &                                     &106.6M   &15.0  &\textbf{97.78}  &84.29     \\
        Patternformer (Res50-ViT$S$)              &                                     &45.2M    &7.0   &97.73  &82.95     \\
        Patternformer (Res50-ViT$B$)              &                                     &109.0M   &15.4  &97.68  &\textbf{84.96}     \\
        
        Patternformer (Efficient-T)                     &                                     &11.9M   &1.6  &97.15  &81.63     \\
        Patternformer (Efficient-S)                     &                                     &29.8M   &3.5  &97.61  &82.33     \\
        Patternformer (Efficient-B)                     &                                     &56.7M   &6.3  &97.57  &83.35     \\
        \bottomrule
    \end{tabular}
\caption{Comparisons with previous results on CIFAR.} 
\label{sota_cifar}
\end{table*}

\begin{table*}[!ht]
\centering
    \begin{tabular}{lcc|c}
        \toprule
        Method & Params & GFLOPs  & ImageNet \\ \hline\hline
        DeiT-Ti \cite{touvron2021training}             &5.7M   & 1.1    &72.2      \\
        ConViT-Ti \cite{d2021convit}          &6.0M   & 1.0    &73.1      \\
        LocalViT-Ti \cite{li2021localvit}        &5.9M   & 1.3    &74.8      \\
        PVT-Ti \cite{wang2021pyramid}             &13.2M   & 1.9    &75.1      \\
        Patternformer (Efficient-T)            &11.9M   &1.6   &\textbf{75.4}     \\
        \bottomrule
    \end{tabular}
\caption{Comparisons with previous results on ImageNet.} 
\label{sota_imagenet}
\end{table*}

\section{Experiments}
We conduct comprehensive experiments to evaluate the effectiveness of our Pattern Transformer, including extensive ablative studies and visualization. 

\textbf{Datasets}. Both large-scale ImageNet dataset and small-scale CIFAR dataset are adopted to evaluate our model. ImageNet dataset consists of 1.28M training images and 50k validation images from 1000 classes. CIFAR-10 and CIFAR-100 datasets consist of 50k training images and 10k validation images, respectively from 10 and 100 classes. Especially, all experiments on CIFAR datasets are trained from scratch.

\textbf{Setting Up}. 
We construct our Pattern Transformer by integrating the fundamental configurations of ResNet and ViT architectures. We primarily draw upon the training recipes in MAE \cite{he2022masked} for stable training. Notably, we opt not to employ color jittering, repeated augmentation, gradient clipping, and layer scaling techniques. Moreover, we distinctly remove the strategies of random erasing and exponential moving average (EMA) in MAE. To capitalize on limited GPU resources, we leverage multiple gradient accumulation steps, enabling the effective large batch size.
Specifically, for the CIFAR dataset, we employ a batch size of 1024 with 8 gradient accumulation iterations on a single NVIDIA GeForce RTX 3090 GPU. For the ImageNet dataset, we utilize a batch size of 4096 with 2 gradient accumulation iterations on four NVIDIA A100 GPUs. Furthermore, we report the final accuracy for 800 and 300 epochs on CIFAR and ImageNet datasets, respectively, And all other experiments on CIFAR datasets are conducted with 400 epochs. More comprehensive details are provided in the supplementary materials.

\subsection{Comparison with Previous Works}
As shown in Table \ref{sota_cifar}, our Pattern Transformer achieves state-of-the-art performance on CIFAR-10 and CIFAR-100 datasets. On CIFAR-100 dataset, Pattern Transformer (Res50-ViT$B$)  attains the best performance with an accuracy of 84.96\%, significantly surpassing the previous best model, CCT-7/3x1 + TokenMixup, which scored 83.57\%. 
Additionally, Pattern Transformer (Res34-ViT$B$) achieves a leading accuracy on CIFAR-10 at 97.78\%, surpassing the existing benchmarks such as NesT-B (97.20\%) and TokenMixup (97.75\%). 

It's worth noting that both CCT and NesT employ the network structures specifically designed for small datasets, especially CCT, which takes low-resolution images of $32 \times 32$ as inputs, significantly enhances computational efficiency. Setting aside these works, Pattern Transformer (Efficient-T) model, despite having only 11.9M parameters and requiring 1.6 GFLOPs, still achieved competitive performance with 97.15\% accuracy on CIFAR-10 and 81.63\% on CIFAR-100. This highlights the efficiency and effectiveness of our proposed method.

\begin{table*}[!ht]
\centering
\scriptsize
    \begin{subtable}[t]{0.33\linewidth}
    \centering
        \begin{tabular}{ccccc}
            Block & Width & Params &GFLOPs & Acc(\%) \\ \midrule
            bottle &8 &90M & 12.0 &77.4 \\
            bottle &16 &91M & 12.3 &79.0 \\
            bottle &32 &96M & 13.1 &81.4 \\
            \textbf{basic} &64 &107M & 15.0 & \cellcolor{gray!15} 82.5 \\
            bottle &64 &109M & 15.4 &82.7 \\
            bottle &96 &129M & 18.6 &82.9 \\
        \end{tabular}
        \caption{Building block and width}
    \end{subtable}
    \begin{subtable}[t]{0.33\linewidth}
    \centering
        \begin{tabular}{cccc}
            Stages &Params &GFLOPs & Acc(\%) \\ \midrule
            {[0,0,0,0]}    & 88M & 11.7  & 63.6 \\
            {[2,0,0,0]}    & 88M & 12.2  & 73.2 \\
            {[2,2,0,0]}    & 87M & 12.4  & 77.2 \\
            {[2,2,2,0]}    & 88M & 12.7  & 80.3 \\
            {[2,2,2,2]}    & 97M & 13.1  & 81.4 \\
            {[3,4,6,3]}    & 107M & 15.0  & \cellcolor{gray!15} 82.5 \\
        \end{tabular}
        \caption{Depth}
    \end{subtable}
    \begin{subtable}[t]{0.33\linewidth}
    \centering
        \begin{tabular}{cccc}
            Stages &Params &GFLOPs & Acc(\%) \\ \midrule
            {[1,4,6,3]}    & 106M & 14.5  & 82.5 \\
            {[3,1,6,3]}    & 106M & 14.3  & 82.5 \\
            {[3,4,1,3]}    & 101M & 13.8  & 82.0 \\
            {[3,4,6,1]}    & 97M  & 14.5 & 82.0 \\
            {[1,1,6,3]}    & 106M & 13.8  & 82.2 \\
            {[3,4,6,3]}    & 107M  & 15.0 & \cellcolor{gray!15} 82.5 \\
        \end{tabular}
        \caption{Stage}
    \end{subtable}
\caption{The impact of ResNet variations in building blocks, width, depth, and stages.}
\label{resnet}
\end{table*}

\begin{table*}[!ht]
\centering
\scriptsize
    \begin{subtable}[t]{0.33\linewidth}
    \centering
        \begin{tabular}{ccccc}
            Tokens & Params &GFLOPs & Acc(\%) \\ \midrule
            8 &107M &4.4  &82.1 \\
            16 &107M &5.1  &82.4 \\
            
            32 &107M & 6.5 &82.6 \\
            64 &107M & 9.3 &82.7 \\
            128 &107M & 15.0 & \cellcolor{gray!15} 82.5 \\
        \end{tabular}
        \caption{Tokens}
    \end{subtable}
    \begin{subtable}[t]{0.33\linewidth}
    \centering
        \begin{tabular}{cccc}
            Embedding &Params &GFLOPs & Acc(\%) \\ \midrule
            192    & 27M & 4.4  & 82.6 \\
            
            384    & 43M & 6.6  & 82.7 \\
            576    & 69M & 10.0  & 83.0 \\
            768 & 107M & 15.0  & \cellcolor{gray!15} 82.5 \\
            960    & 154M & 21.2  & 82.4 \\
        \end{tabular}
        \caption{Embedding}
    \end{subtable}
    \begin{subtable}[t]{0.33\linewidth}
    \centering
        \begin{tabular}{cccc}
            Depth &Params &GFLOPs & Acc(\%) \\ \midrule
            
            3    & 43M & 6.5  & 82.4 \\
            6    & 64M & 9.3  & 82.6 \\
            9    & 85M & 12.1  & 82.9 \\
            12    & 107M  & 15.0 & \cellcolor{gray!15} 82.5 \\
        \end{tabular}
        \caption{Depth}
    \end{subtable}
    \begin{subtable}[t]{0.33\linewidth}
    \centering
        \begin{tabular}{cccc}
            Heads &Params &GFLOPs & Acc(\%) \\ \midrule
            1    & 107M & 15.0  & 82.4 \\
            
            3    & 107M & 15.0  & 83.1 \\
            6    & 107M & 15.0  & 83.0 \\
            8    & 107M & 15.0  & 82.5 \\
            12    & 107M  & 15.0 & \cellcolor{gray!15} 82.5 \\
        \end{tabular}
        \caption{Heads}
    \end{subtable}
    \begin{subtable}[t]{0.33\linewidth}
    \centering
        \begin{tabular}{cccc}
            Mlp ratio &Params &GFLOPs & Acc(\%) \\ \midrule
            1    & 64M & 9.5  & 82.5 \\
            2    & 78M & 11.3  & 83.1 \\
            3    & 92M & 13.1  & 83.0 \\
            4    & 107M  & 15.0 & \cellcolor{gray!15} 82.5 \\
        \end{tabular}
        \caption{Mlp ratio}
    \end{subtable}
\caption{The impact of Transformer variations in token number, embedding dimension, depth, attention heads, and MLP ratio.}
\label{vit}
\end{table*}

Table \ref{sota_imagenet} gives the results of our proposed Pattern Transformer with several previous works on the ImageNet dataset. The Efficient-T variant of Pattern Transformer, with 11.9M parameters and 1.6 GFLOPs, achieves the best performance among the compared models with similar complexity. It outperforms DeiT-Ti, ConViT-Ti, LocalViT-Ti, and PVT-Ti by 3.2\%, 2.3\%, 0.6\%, and 0.3\% respectively. Notably, it achieves this superior performance with fewer parameters and computational complexity than PVT-Ti.

\subsection{Ablation Study}
We conduct in-depth ablation studies on numerous variants of the proposed Pattern Transformer, aiming at analyzing the influence of different architectural choices on the model performance. 
Our goal is to identify the key factors that contribute to improved accuracy, while concurrently considering the corresponding model parameters and computational complexity. We observed that an efficient Pattern Transformer relies on a combination of Heavy ResNet and Light Transformer. We set the Res34-ViT$B$ architecture in Table \ref{model} as the baseline, distinctly denoted by a grey block, to facilitate better comparison.

\subsubsection{Heavy ResNet}
ResNet captures all potential patterns within image content and we analyze the impact of ResNet variations in building blocks, width, depth, and different stages as discussed in Section \ref{heavy}. 
As illustrated in Table \ref{resnet}. we found that the model's accuracy incrementally improves as the width of the building block expands. However, the increase in accuracy is accompanied by a growth in both model parameters and computational complexity. Similar to width, a deeper network, such as the stage architecture [3,4,6,3], enhances accuracy by capturing more complex patterns. We also explored the imports of different stage configurations. While reducing the building blocks in the first two stages maintained the original accuracy, reductions in the final two stages resulted in accuracy losses. This suggests that the latter stages play a more significant role in feature extraction. Overall, our ablation study indicates that the choice of building block, its width, the model's depth, and stage configuration are all crucial to performance. Continual enhancement of these factors leads to consistent accuracy improvement, and we did not observe a saturation trend. However, this significantly escalates model parameters and computational complexity.
A configuration using the ``bottle'' block with a width of 96 and a stage structure of [1,1,6,3] provides a good balance between accuracy, model complexity, and computational costs.




\subsubsection{Light Transformer}
Transformer models long-range dependencies among patterns and we examine various aspects of Transformer variations in token number, embedding dimension, network depth, attention heads, and MLP ratio as discussed in Section \ref{heavy}. 
The results are presented in Table \ref{vit}.
We observe a general trend of increasing accuracy with an increase in tokens, with the highest accuracy of 82.7\% attained at 64 tokens. However, beyond this point, the accuracy decreases slightly, suggesting that there is an optimal token size for this task, after which the model performance may degrade due to overfitting or increased complexity. Similar to token number, embedding dimension, network depth, attention heads, and MLP ratio all play crucial roles in the model's performance. The setting of 64 pattern tokens with an embedding dimension of 576 in a 6 layers of Transformer with 6 heads, 2 MLP ratio offers a good balance between accuracy, model complexity, and computational costs.

\subsection{Discussions}

\begin{table}[!t]
\centering
    \begin{tabular}{lc|c}
        \toprule
        Method & \begin{tabular}[c]{@{}c@{}}Tokenizer\\ Layer\end{tabular}   & CIFAR100 \\ \hline\hline
        Our Pattern Transformer             &     &      \\ \hline
        ViT stem             & 1    &54.9      \\ 
        Basic [0,0,0,0]            & 2    &63.6      \\ 
        Basic [2,0,0,0]            & 6    &73.2      \\ 
        Basic [3,4,6,3]            & 34    &82.9      \\
        Bottle [3,4,6,3]           & 50    &83.4      \\ 
        Bottle [3,4,6,4]           & 53    &83.6      \\ \hline \hline
        Our Patch Transformer            &  &     \\ \hline
        Bottle [3,4,6,3]            & 50    &83.7      \\ 
        \bottomrule
    \end{tabular}
\caption{In-depth experiments of Network Depth.} 
\label{complex}
\end{table}

\textbf{Why does Pattern Transformer require a more complex tokenizer}? 

In conventional Vision Transformers, the size of the patches plays a pivotal role in balancing the speed-accuracy tradeoff, with smaller patches leading to higher accuracy but at a higher computational cost. For instance, the recent FlexiViT has found that ViT-B with a patch size of $8 \times 8$ achieves an accuracy of 85.6\% on ImageNet1k with 156 GFLOPs, whereas ViT-B with a patch size of $32 \times 32$ only attains an accuracy of 79.1\% with 8.6 GFLOPs. This is due to the fact that as the patch size increases, the transformer struggles to optimize local information and is forced to reduce the sequence length. However, this also underscores the effectiveness of our Pattern Transformer in a way, as it can be simply seen as having a patch size of $224 \times 224$, although it's not entirely equivalent.

The Pattern Transformer maximizes the patch size, thus exacerbating the negative impacts mentioned above. As depicted in the figure \ref{complex}, Pattern Transformer with one-layer tokenizer (similar to ViT stem) merely achieves an accuracy of 54.9\%. To counter this, we necessitate a more complex tokenizer to extract higher-level semantic information. The Pattern Transformer with a 50-layer ResNet50 tokenizer can achieve an accuracy of 83.4\%. By further increasing the fourth stage of the bottleneck block, an accuracy of 83.6\% can be achieved. Therefore, this complexity is integral to the Pattern Transformer's ability to excel in image classification tasks.

As a byproduct, our Flexsible Patch Transformer, before being fed into the transformer, completely scrambles the original patch sequence space through an extra linear transformation (brutally converting 49 patches into 128 patches). This process entirely loses the interpretability, yet it achieves the best performance with an accuracy of 83.7\%.

\subsection{What exactly is the manifestation of Visual Words?}
\begin{figure}[t]
\centering
\includegraphics[width=\columnwidth]{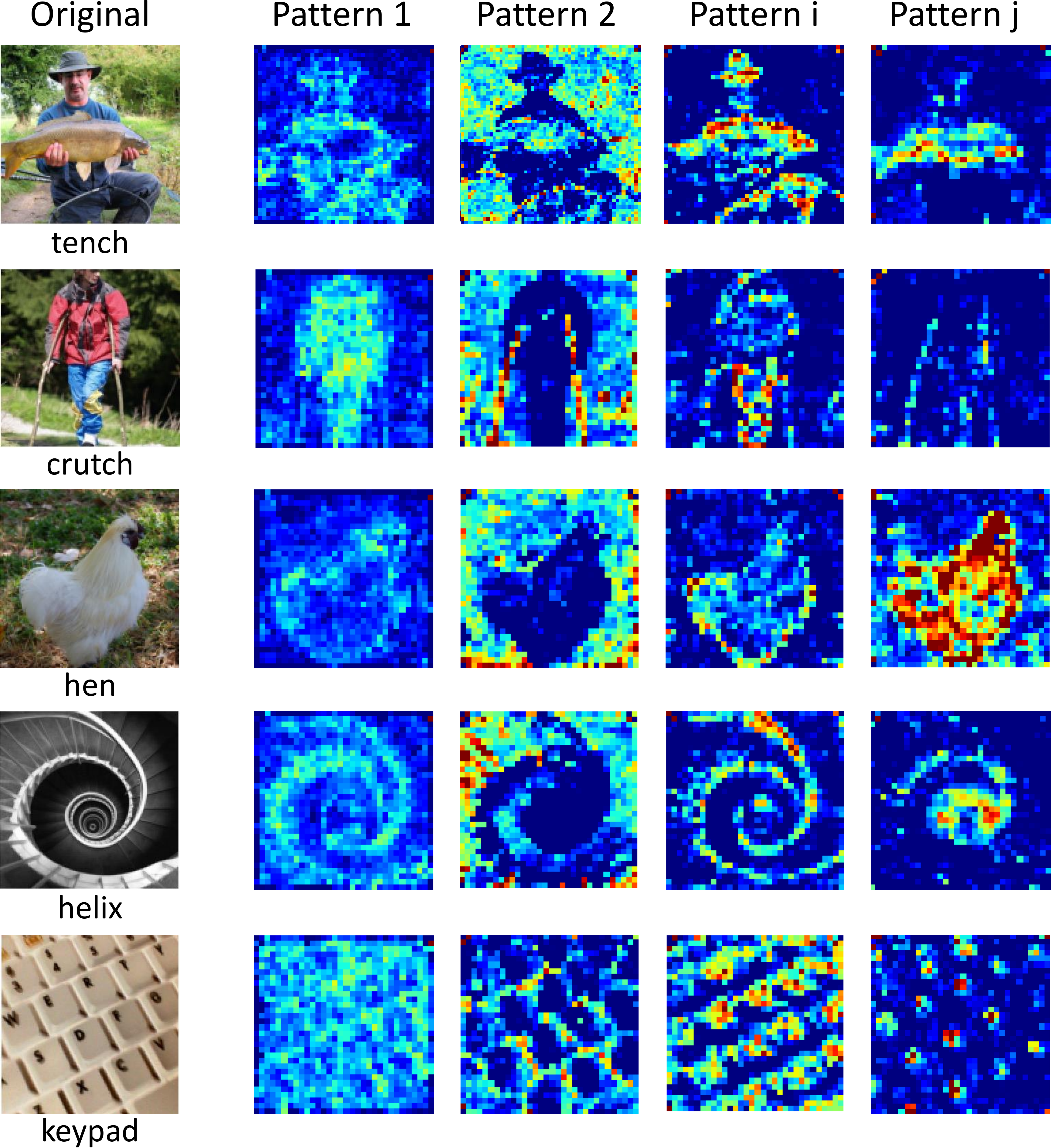} 
\caption{The visualization of Visual Words.}
\label{intro}
\end{figure}

As we emphasized earlier, each pattern captures its local region of interest, thereby preserving its intrinsic structural and semantic information. To illustrate this, we visualized the Pattern Transformer (Res50-ViTB) architecture, demonstrating the visual tokens before they are fed into the transformer. The Res50-ViTB architecture generates a total of 128 patterns. We observed that certain patterns, such as 1st, 17th, 35th, and 102nd, consistently capture the foreground information of the object, while certain patterns like 2nd, 14th, 44th ,123th consistently capture the background information. As depicted in Figure 4, the first two columns invariably output the feature maps of the first and second patterns, while the third and fourth columns are selected randomly. We noted that the first pattern always prioritizes capturing the foreground related to humans, such as the low response of fish overlapping with the human body in the ``tench''. Conversely, the second pattern always prioritizes focusing on the interesting background, particularly the background interacting with humans, such as the cane in the ``crutch''. Furthermore, for pattern j in the fourth column, we visualized interesting visual tokens, such as the fish in ``tench'' and the letters on ``keypad''. These visualizations encompass comprehensive semantic information, further substantiating our claim, the efficacy of the Pattern Transformer in preserving the intrinsic structural and semantic information.

\section{Conclusion}
Pattern Transformer provides a novel solution to the challenge of converting images into sequence inputs for Vision Transformers. 
Traditional patchification disrupts the cohesion of structural and semantic details within images, constraining the potential of Transformer-based models.
Pattern Transformer addresses these issues by dynamically converting images into pattern sequences. Leveraging Convolutional Neural Networks (CNNs), we extract localized patterns, assigning individual channels to specific patterns. These patterns are treated as visual words, seamlessly integrated into the Transformer network known for capturing global contextual relationships. By merging local patterns with global features, we effectively synergize CNNs and Transformers.
Future work includes investigating the effectiveness of Pattern Transformer on other visual tasks such as detection and segmentation.


\bibliography{aaai24}

\clearpage
\appendix
\section{More Discussions}
\subsection{Compact Visual Words}
Images inherently possess a high level of redundancy. Manual partitioning of images into a fixed number of patches often produces redundant and uninformative patches for the target class. These redundant patches not only fail to provide beneficial information but also impose a computational burden on expensive Transformer. 
\begin{figure}[!ht]
\centering
\includegraphics[width=0.36\textwidth]{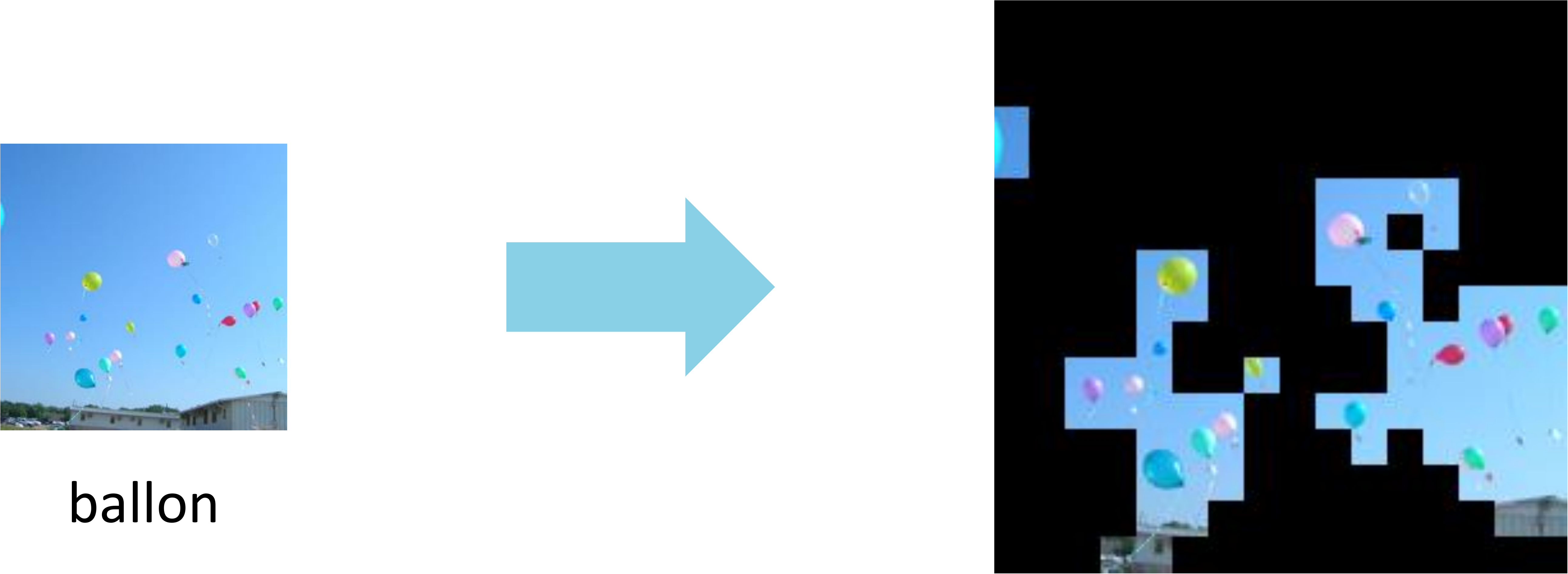}
\caption{Vision Transformer contains redundant patches.}
\label{bu_1_1_c}
\end{figure}
As exemplified in Figure \ref{bu_1_1_c}, merely 68 out of 256 patches furnish discriminative information for the "balloon" class, whereas the remaining 188 patches predominantly contain superfluous details, such as sky, buildings, trees, etc. All these redundant patches partake in the subsequent Transformer computations, thereby exacerbating the redundancy. This amplification of redundancy becomes particularly pronounced when objects are situated at a considerable distance.
\begin{figure}[!ht]
  \centering
  \begin{subfigure}{0.46\textwidth}
        \centering
        \includegraphics[width=0.86\textwidth]{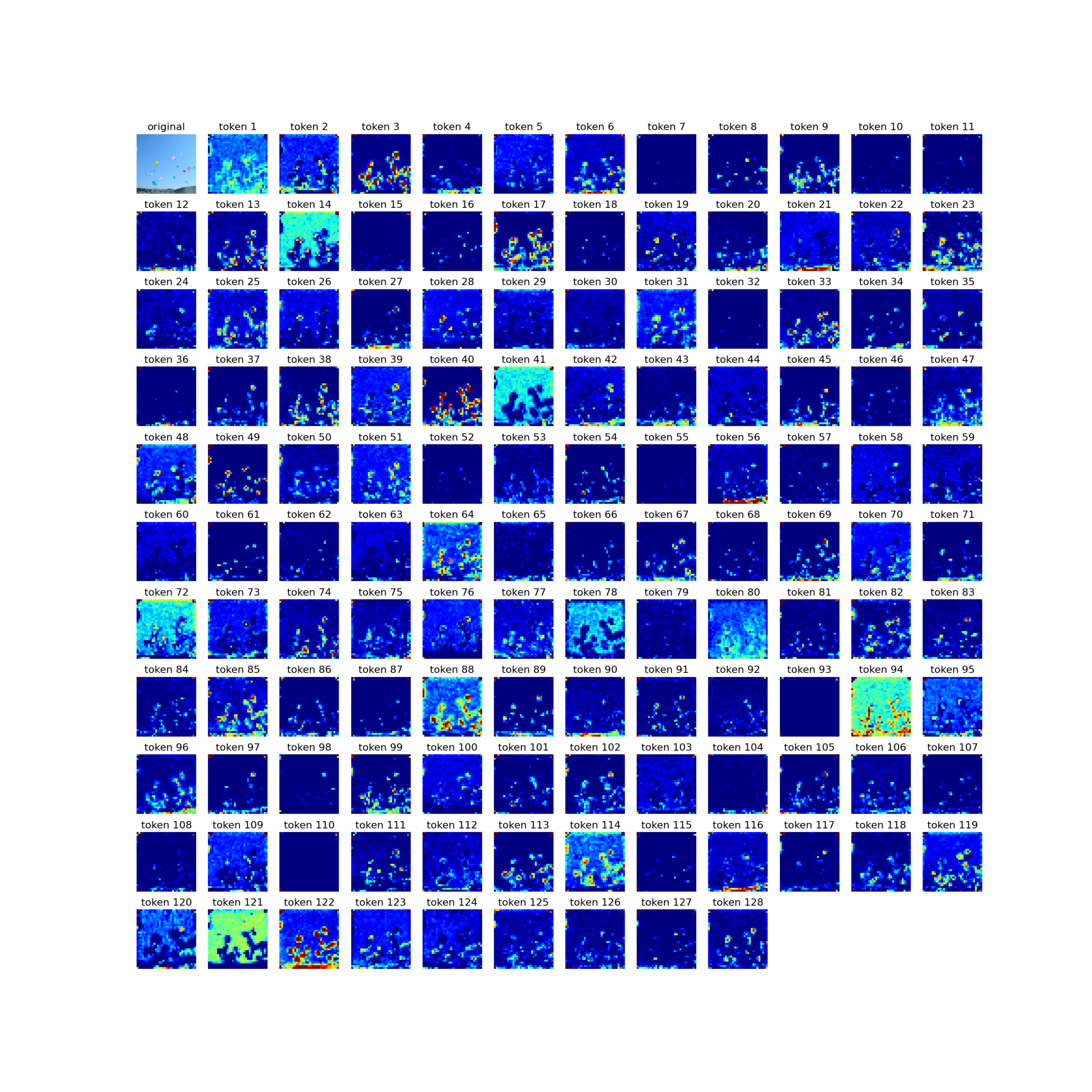}
        \caption{128 patterns in our Res$50$ - ViT$B$ architecture.}
        \label{bu_1_2_c}
    \end{subfigure}
    \begin{subfigure}{0.46\textwidth}
        \centering
        \includegraphics[width=0.86\textwidth]{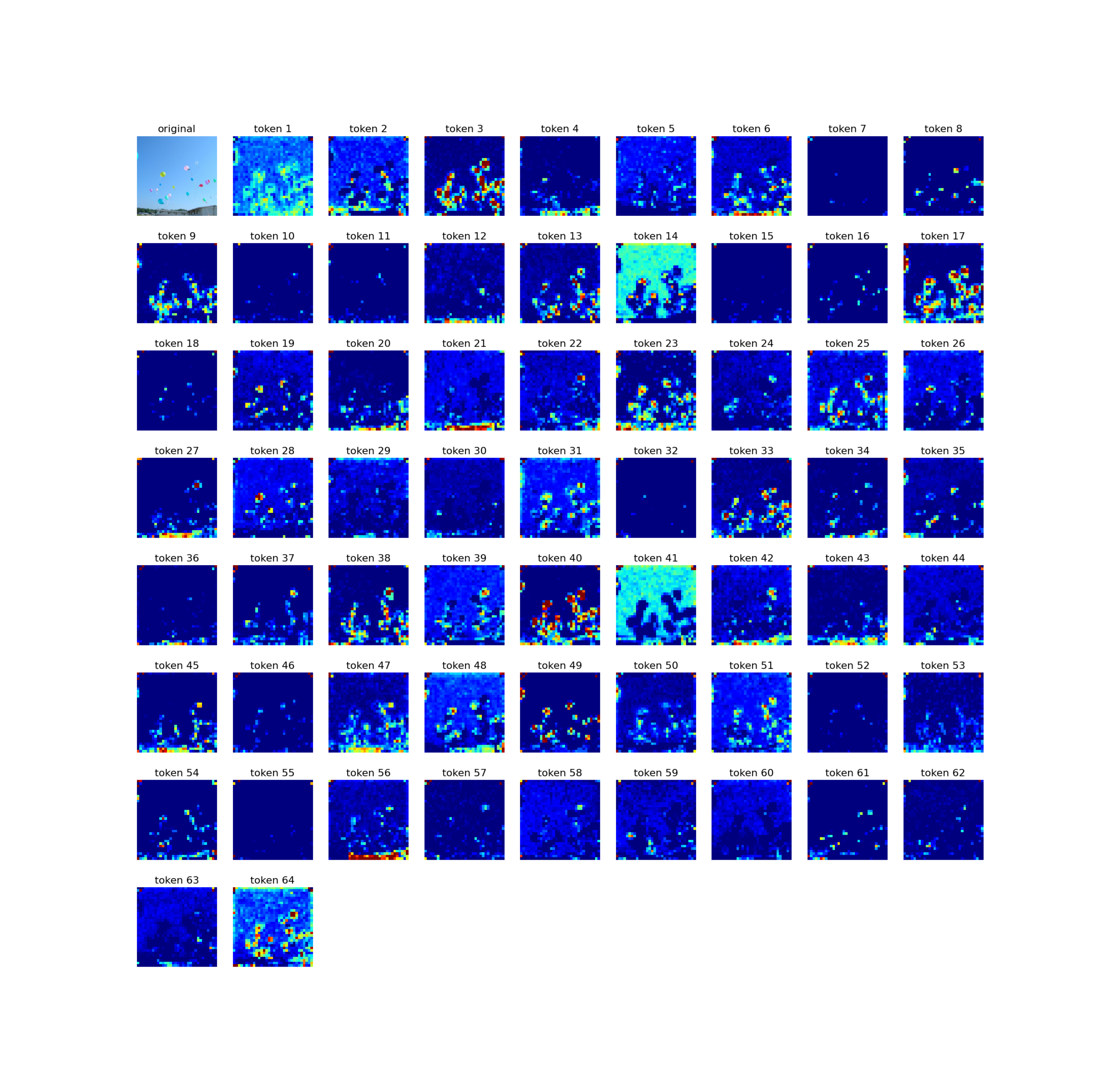}
        \caption{64 patterns in our Efficient-$B$ architecture.}
        \label{bu_1_3_c}
    \end{subfigure}
  \caption{The visualization of compact visual tokens.}
  \label{bu_1}
\end{figure}

Pattern Transformer converts the image into a sequence of patterns (visual tokens), where each pattern captures the local region of interest. These patterns are autonomously optimized by the network, progressively extracting features. As depicted in Figures \ref{bu_1_2_c} and \ref{bu_1_3_c}, we have visualized 128 various patterns in Res$50$ - ViT$B$ and 64 patterns in Efficient-$B$ respectively. The majority of these patterns capture the balloon area, with varying responses focusing on different balloon aspects. A handful of these features exhibit low response activation or may concentrate on the background area, a phenomenon less frequent in Efficient architecture. Anyway, compared to the conventional Transformer, our Pattern Transformer is more compact and efficient.

\begin{figure*}[!htb]
\centering
\vspace{-5pt}
\includegraphics[width=0.62\textwidth]{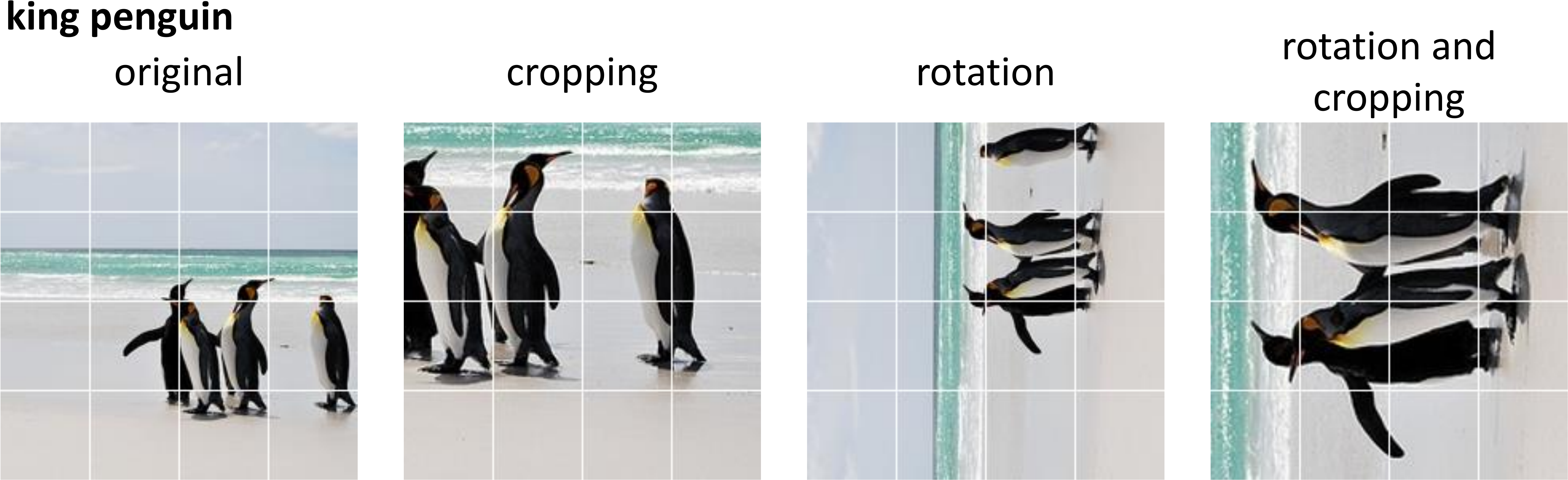}
\caption{Semantic inconsistency in Vision Transformer.}
\label{bu_2_1_c}
\end{figure*}

\begin{figure*}[!h]
  \centering
  \begin{subfigure}{0.46\textwidth}
        \centering
        \includegraphics[width=0.86\textwidth]{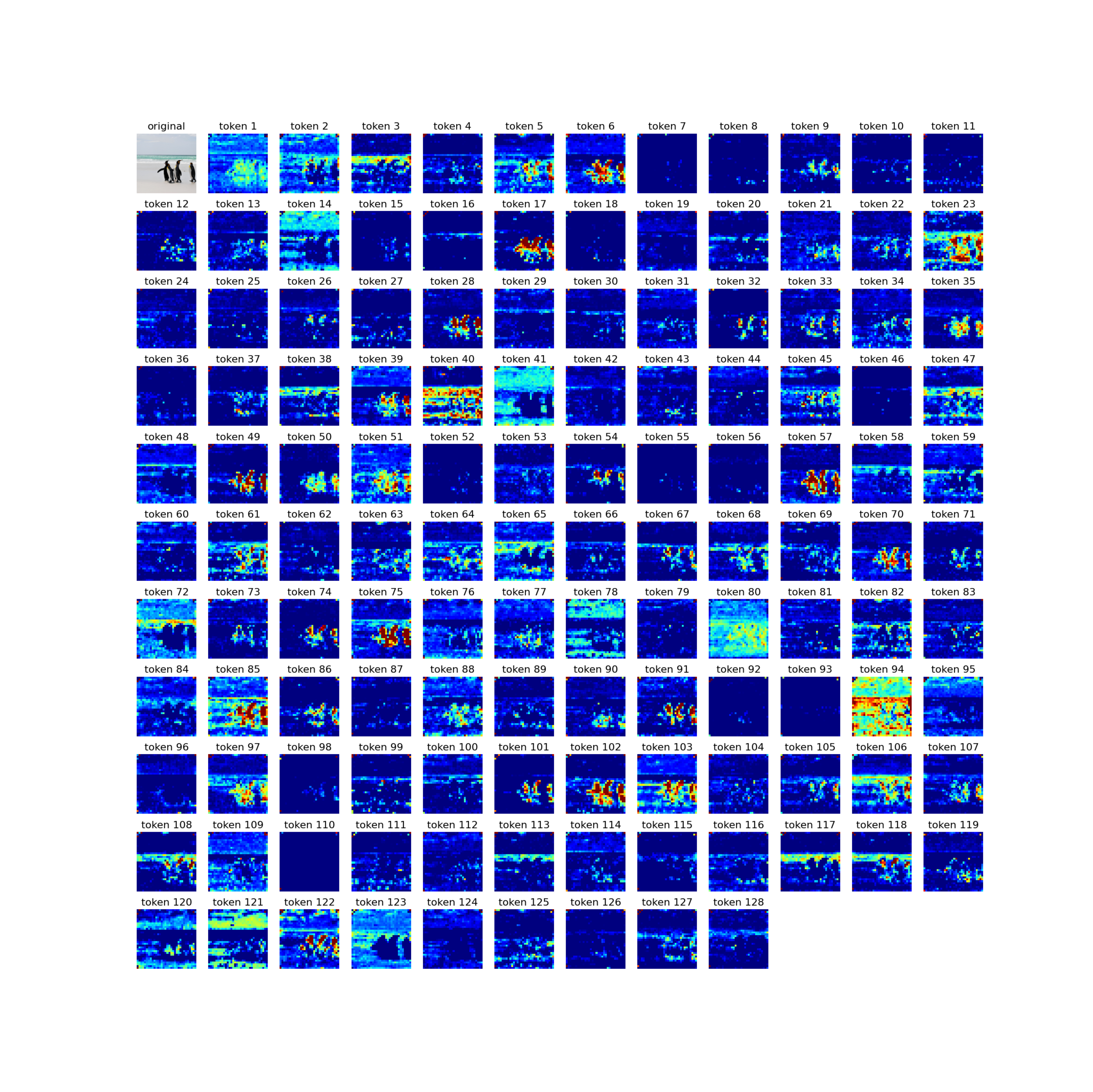}
        \caption{original}
        \label{bu_2_2_c}
    \end{subfigure}
    \begin{subfigure}{0.46\textwidth}
        \centering
        \includegraphics[width=0.86\textwidth]{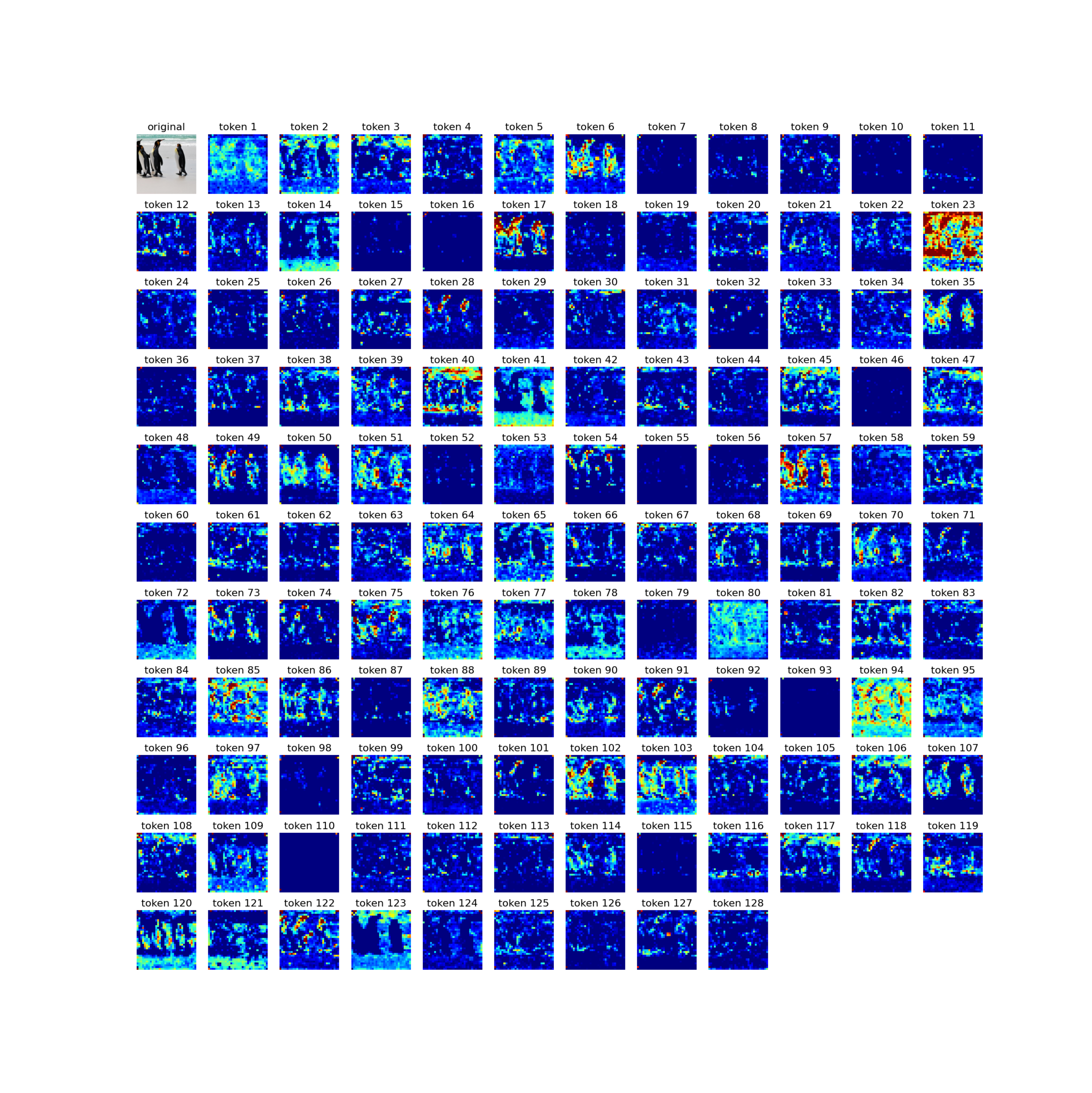}
        \caption{cropping}
        \label{bu_2_3_c}
    \end{subfigure}
    \\
  \begin{subfigure}{0.46\textwidth}
        \centering
        \includegraphics[width=0.86\textwidth]{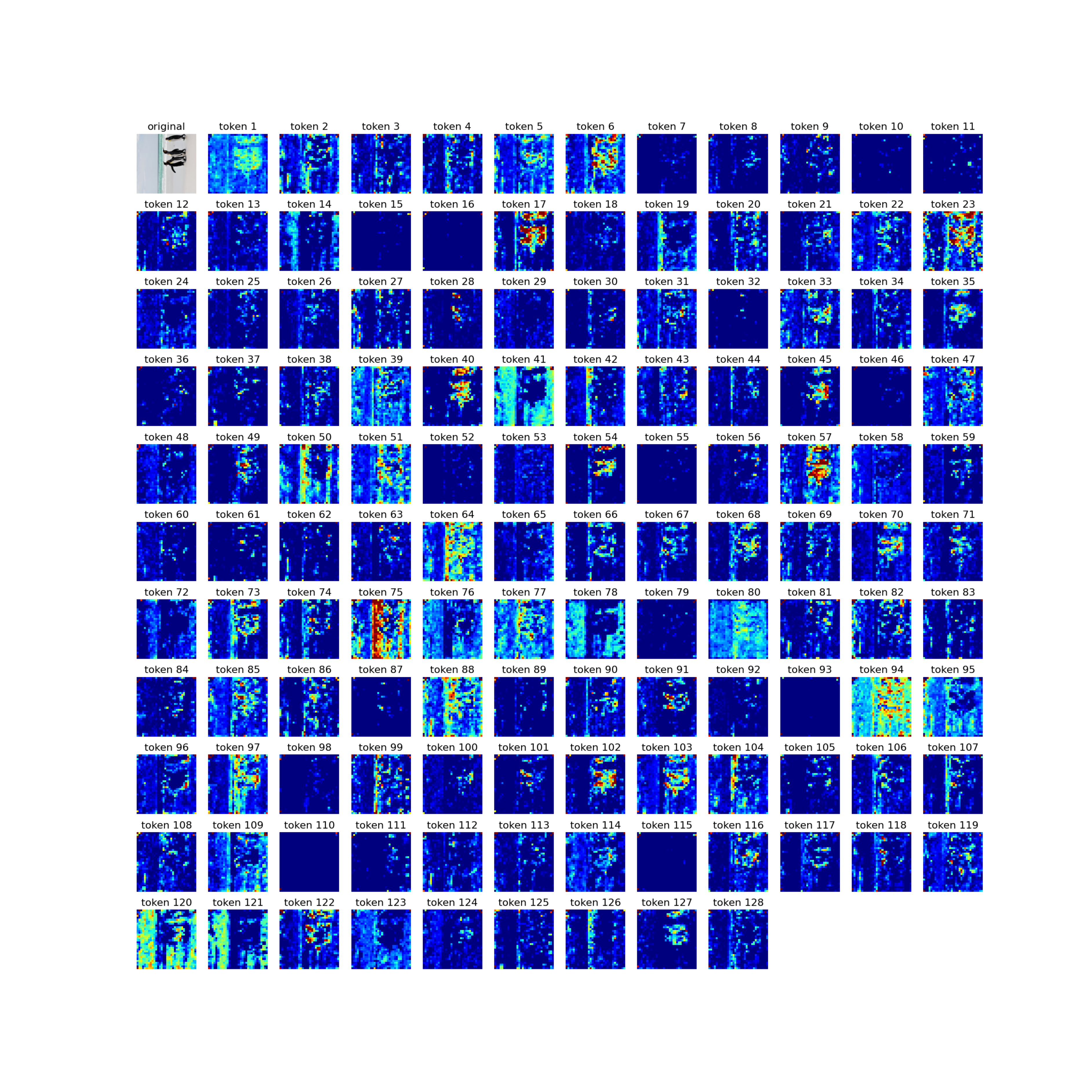}
        \caption{rotation}
        \label{bu_2_4_c}
    \end{subfigure}
    \begin{subfigure}{0.46\textwidth}
        \centering
        \includegraphics[width=0.86\textwidth]{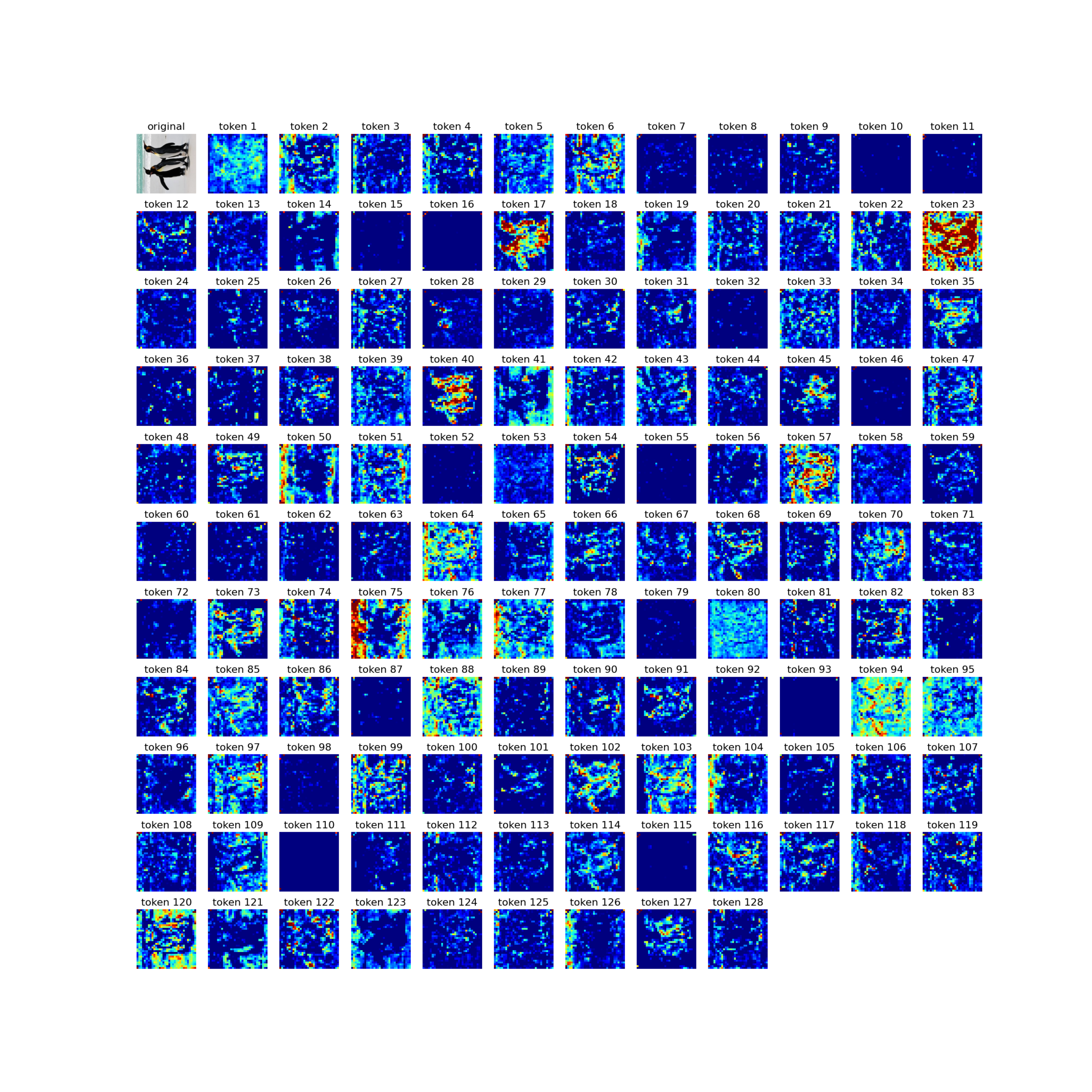}
        \caption{rotation and
cropping}
        \label{bu_2_5_c}
    \end{subfigure}
  \caption{The visualization of semantically consistent visual tokens.}
  \label{bu_2}
  \vspace{-5pt}
\end{figure*}

\subsection{Semantic Consistency}
Images, especially complex ones, often depict identical objects with different geometric variations caused by rotation and scale. Manually dividing these images into rigid patches based on experiential knowledge results in inconsistent semantic information. As exemplified in Figure \ref{bu_2_1_c}, considerable changes occur in the patch sequence of the same image after simple cropping or rotation operations.

Convolutional neural networks excel at capturing intricate patterns, and our Pattern Transformers naturally inherit its ability to process shift-invariant patterns in images. Figure \ref{bu_2} visualizes patterns (visual tokens) under cropping, rotation, and cropping with rotation, observing consistent semantic information. For instance, token 1 invariably prioritizes capturing the foreground related to specific information, such as penguins. Conversely, token 2 consistently focuses on the interesting background, like coastlines. These observations indicate that our Pattern Transformer maintains excellent semantic consistency.

\subsection{Preservation of Complete Structure}
Traditional Transformers inherently depend on sequential input, necessitating the manual partitioning of images into patch sequences, which disrupts the inherent image structure. 
In contrast, Pattern Transformer converts the image into a sequence of patterns. Thanks to its global receptive field, it can capture all potential image content concepts, thereby preserving the complete structural information.

\subsection{Flexsible Sequence Length}
The sequence length of the traditional Transformer is determined by the image resolution and patch size, which limits the model's potential and flexibility. Our Pattern Transformer aligns channels with the sequence length in Transformer, enabling the computation of self-attention among patterns of a flexible value, which provides a trade-off between speed and accuracy.

\clearpage
\subsection{Limitations}
Revisiting the experiment results outlined in Section 4, our Pattern Transformer (Res50-ViT$B$) demonstrates the best performance, achieving an impressive accuracy of 84.96\%, significantly outperforming prior works by a large margin. Notably, the Pattern Transformer (Efficient-T), despite its compact architecture with merely 11.9M parameters and requiring 1.6 GFLOPs, still delivers  a competitive performance of 81.63\%. However, we noticed generally low performance across pure transformers. For instance, Swin-B, despite boasting 88M parameters, only reaches an accuracy of 78.5\%, while DeiT-B yields an even  lower performance, with an accuracy of just 70.5\%. These observations prompt us to delve deeper into the underlying causes of these phenomena.

As illustrated in Figure \ref{limit}, we retrained the ViT-B and ResNet50 on CIFAR-100 under identical settings, yielding intriguing results. Specifically, our reimplementation ResNet50 achieved an accuracy of 81.9\%. This sheds light on three key phenomena:
\begin{itemize}
    \item The benefits derived from the convolutions play a pivotal role.
    \item By utilizing patterns from convolutional networks as visual tokens and further deploying transformers to model global information, we can capture more comprehensive features. Compared to ResNet50, the Pattern Transformer (Res50-ViT$B$) still offers a performance improvement of 1.5
    \item The conventional convolution, when combined with novel training strategies, still maintains a significant advantage on CIFAR-100. We have identified and provided a superior training recipe for ResNet50.
\end{itemize}
Moreover, this further corroborates our assertion in Section 4 that the performance would markedly deteriorate when employing a lightweight tokenizer. This implies that the limitations of the Pattern Transformer lie in its need for robust and efficient pattern extraction.

\begin{table}[!ht]
\centering
    \begin{tabular}{l|c}
        \toprule
        Method &  CIFAR100 \\ \hline\hline
        ViT-B \cite{zhang2022nested} &70.5      \\ 
        ResNet50 \cite{choi2022tokenmixup} & 61.7 \\ \hline
        ViT-B  &75.2      \\ 
        ResNet50  & 81.9 \\ 
        Patternformer (Res50-ViT$B$)             & 83.4      \\ 
        \bottomrule
    \end{tabular}
\caption{Ablation Studies on two factions} 
\label{limit}
\end{table}

\subsection{More Visualization}
\begin{figure*}[]
\centering
\includegraphics[width=0.98\textwidth]{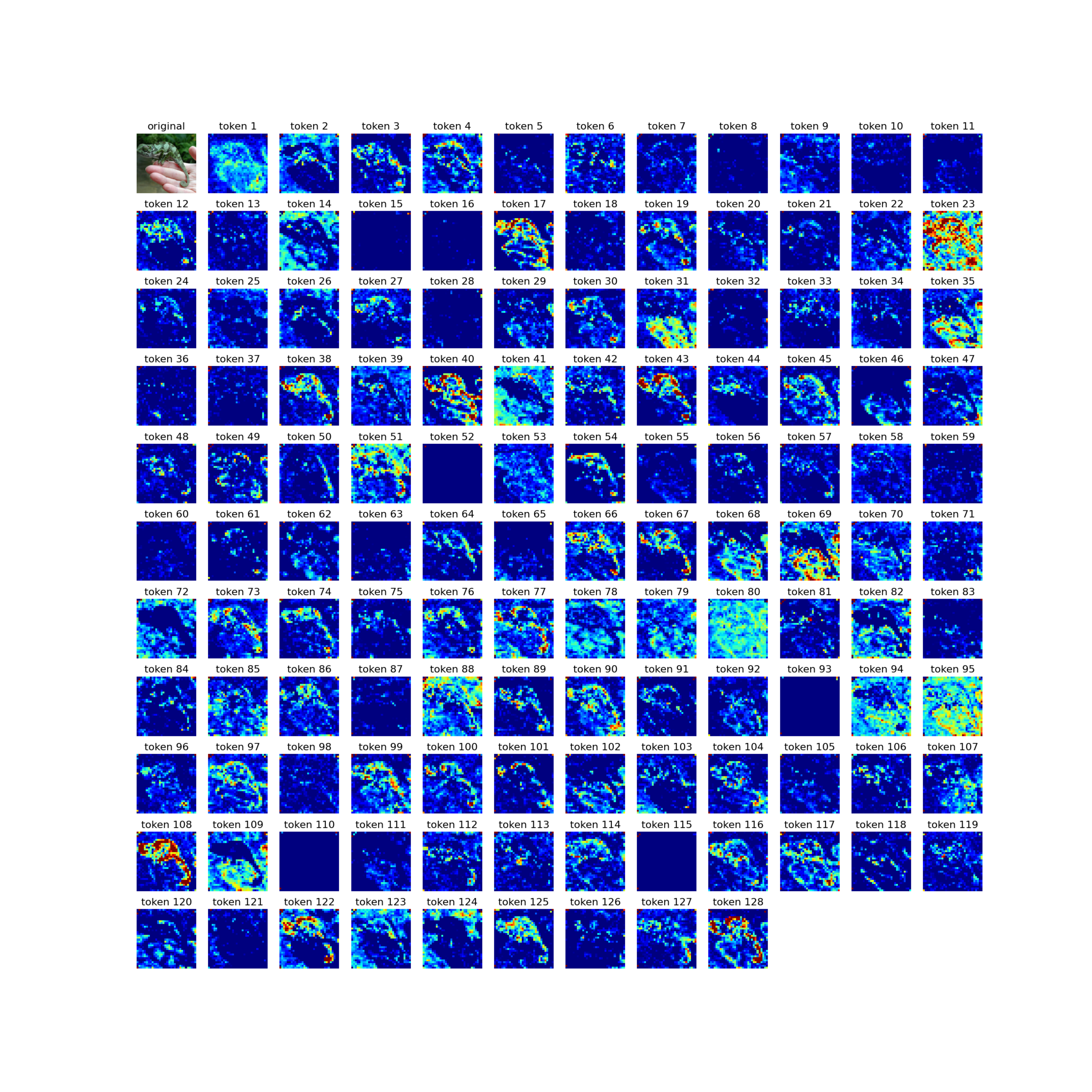}
\caption{The visualization of visual tokens for ``African chameleon''.}
\label{bu_3_1}
\end{figure*}

\begin{figure*}[]
\centering
\includegraphics[width=0.98\textwidth]{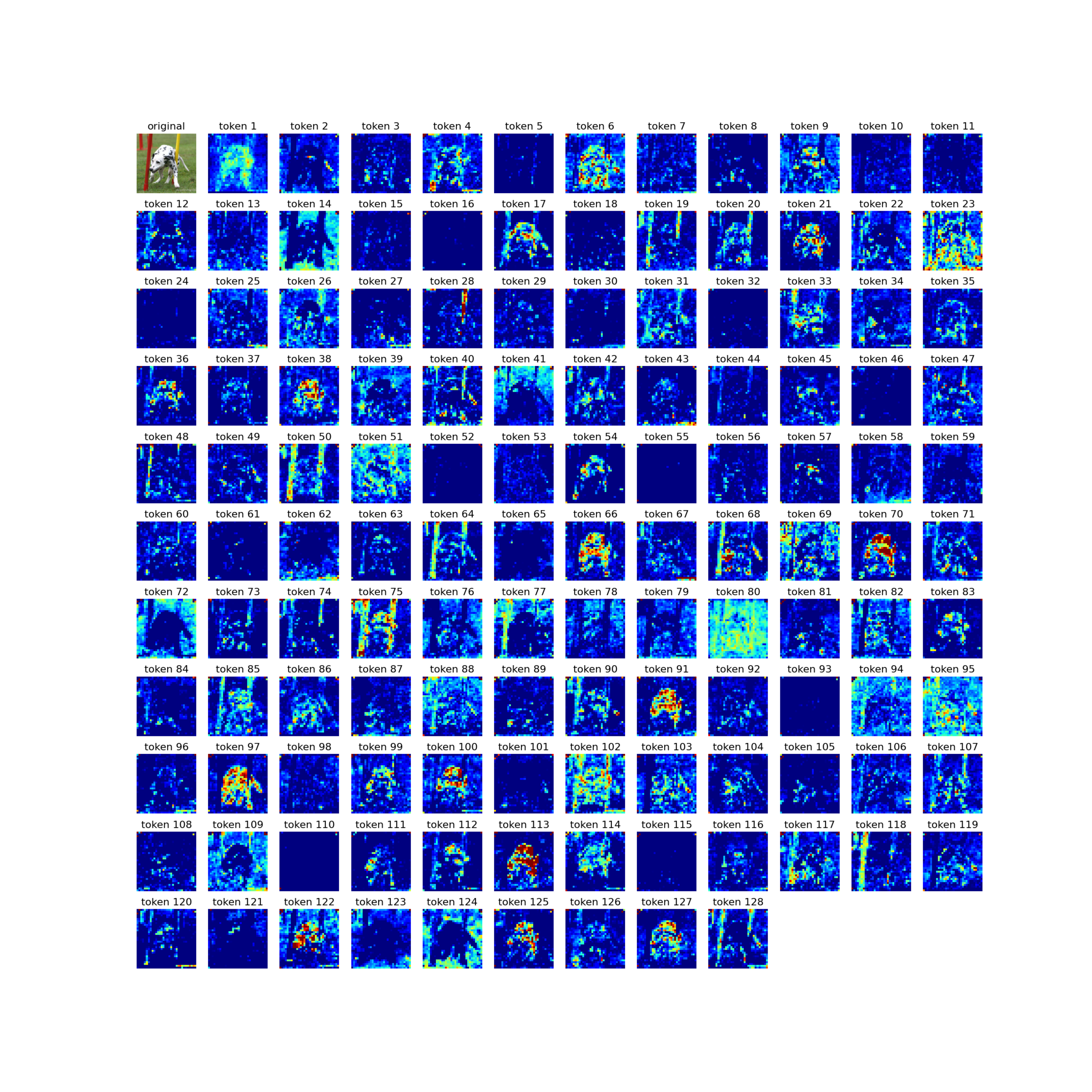}
\caption{The visualization of visual tokens for ``dalmatian''.}
\label{bu_3_2}
\end{figure*}

\begin{figure*}[]
\centering
\includegraphics[width=0.98\textwidth]{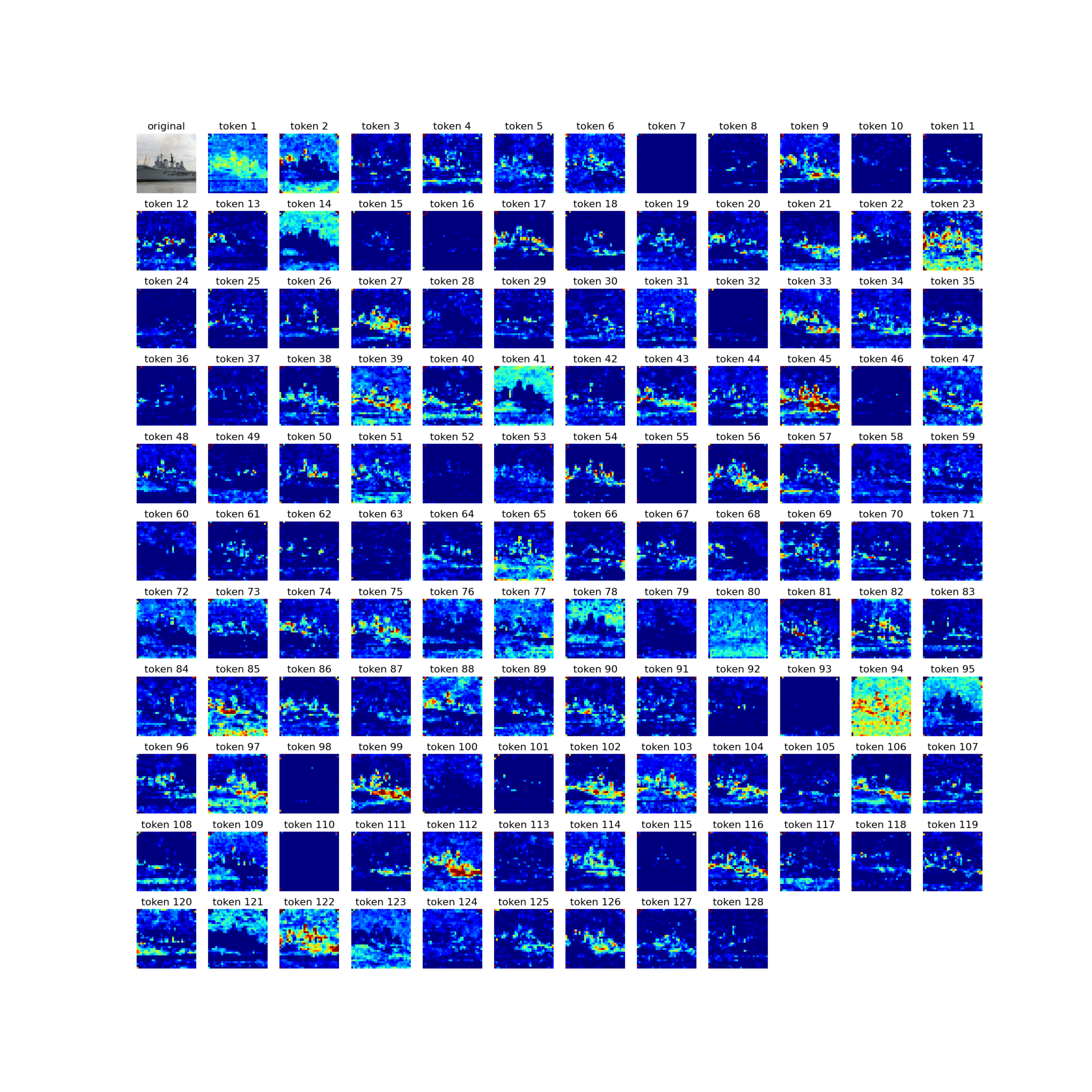}
\caption{The visualization of visual tokens for ``aircraft carrier''.}
\label{bu_3_3}
\end{figure*}

\begin{figure*}[]
\centering
\includegraphics[width=0.98\textwidth]{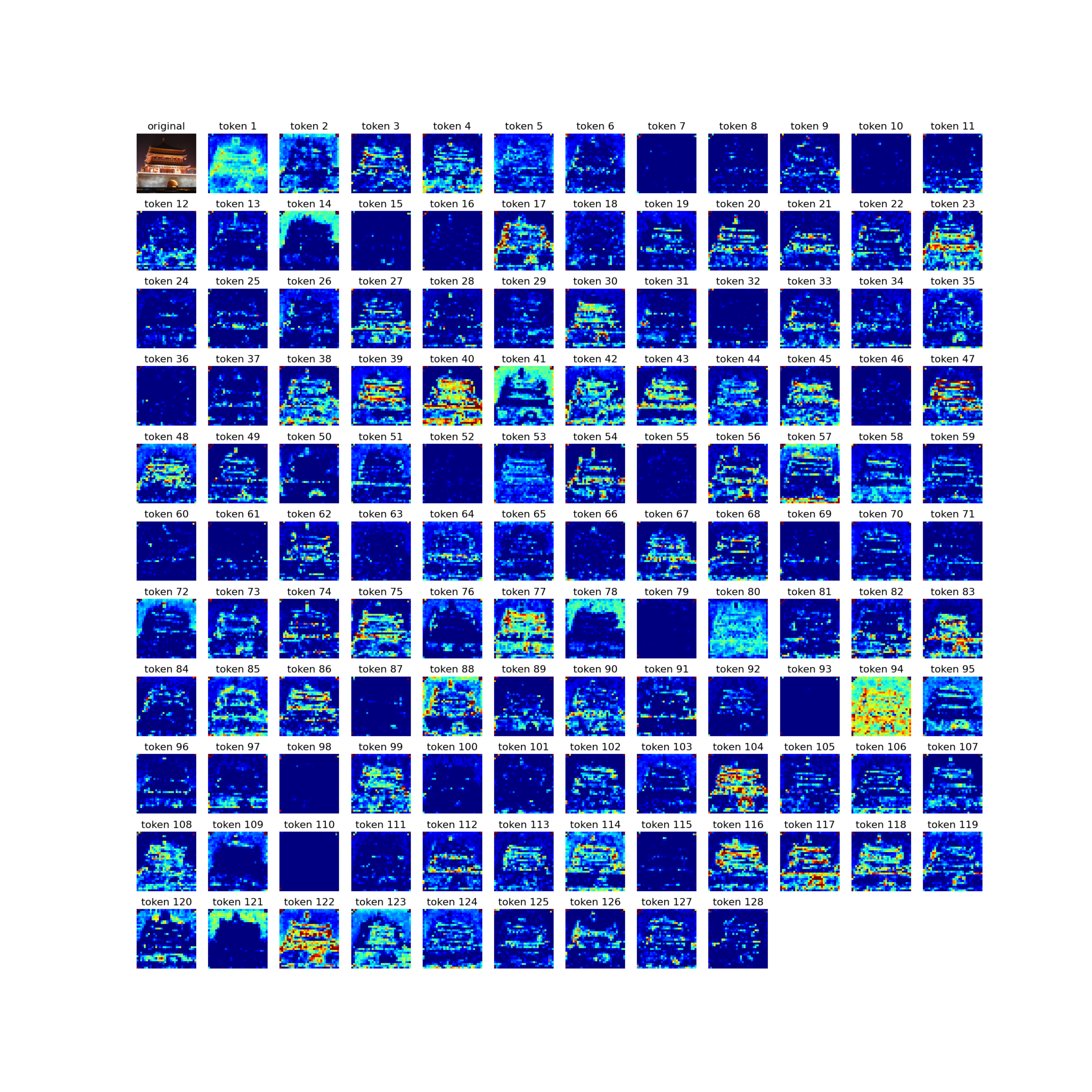}
\caption{The visualization of visual tokens for ``bell cote''.}
\label{bu_3_4}
\end{figure*}

We provide more visualization results at the end of our paper as shown in Figure \ref{bu_3_1} - \ref{bu_3_4}. These results further substantiate the efficacy of our Pattern Transformer and visually offer a more comprehensive understanding of our research.

\section{Experiment Details}
Further details on our experimental settings are provided in Table \ref{setting}. In contrast to recent studies, which often differ in regularization and augmentation usage, frequently intensify these with larger model sizes to prevent overfitting and enhance accuracy, our method maintains a consistent training recipe across all variants and delivers robust performance without extensive hyper-parameter search.
\begin{table}[!ht]
\centering
    \begin{tabular}{l|l}
            Config      & Value     \\ \hline
     Resolution (Train and Test)     &   224 $\times$ 224   \\
     Optimizer      &  AdamW\\
     Batch size      &  1024(C), 4096(I)\\
     Base learning rate      & 2e-4(C), 1e-4(I) \\
     Learning rate schedule     &  cosine \\
     layer-wise lr decay    &  0.65 \\
     Weight decay   &  0.3\\
     Momentum       &  $\beta_1, \beta_2$ = 0.9, 0.95\\
     Training epochs         &  800(C), 300(I)\\
     Warmup epochs  & 5(C), 20(I) \\ \hline

     Label smoothing & 0.1 \\
     Dropout        & \textcolor[rgb]{0.6,0.6,0.6}{\ding{55}} \\
     Drop path       & 0.1 \\
     Repeated Aug   & \textcolor[rgb]{0.6,0.6,0.6}{\ding{55}} \\
     Gradient Clip  & \textcolor[rgb]{0.6,0.6,0.6}{\ding{55}} \\ \hline
     
     Rand Augment   & 9 / 0.5 \\
     
     Mixup              & 0.8 \\
     Cutmix         & 1.0 \\
     Erasing prob   & \textcolor[rgb]{0.6,0.6,0.6}{\ding{55}} \\
     ColorJitter    & \textcolor[rgb]{0.6,0.6,0.6}{\ding{55}} \\
     PCA lighting    & \textcolor[rgb]{0.6,0.6,0.6}{\ding{55}} \\

     Loss           & CE \\ \hline
     LayerScale     & \textcolor[rgb]{0.6,0.6,0.6}{\ding{55}} \\
     SWA            & \textcolor[rgb]{0.6,0.6,0.6}{\ding{55}} \\
     EMA            & \textcolor[rgb]{0.6,0.6,0.6}{\ding{55}} \\

    \end{tabular}
\caption{Training settings. ``C'' represents CIFAR-10 and CIFAR-100 datasets, while ``I'' represents ImageNet dataset.} 
\label{setting}
\end{table}

\begin{table}[!ht]
\centering
    \begin{tabular}{l|c}
        \toprule
        Strategies &  CIFAR100 \\ \hline\hline
        Patternformer (Res50-ViT$B$)             & 83.4      \\ \hline
        w.o. smooth                       &82.7      \\ 
        w.o. cutmix                       &81.3      \\ 
        w.o. mixup                       &82.3      \\ 
        w.o. drop path                       &83.3      \\ 
        \bottomrule
    \end{tabular}
\caption{Ablation Studies on Training Methods} 
\label{ab}
\end{table}

Notably, we adopt the linear $lr$ scaling rule: $lr$ = base $lr$ × batchsize / 256. Moreover, we employ global average pooling features for the final classification, rather than the class token. Further ablation experiments regarding the training strategies employed above on CIFAR-100 are detailed in Table \ref{ab}. All cifar ablation experiments are conducted over 400 epochs.


\end{document}